\documentclass[letterpaper, 10 pt, conference]{ieeeconf}  

\IEEEoverridecommandlockouts                              
\usepackage{graphicx}    
\usepackage{epsfig}
\usepackage{float}
\usepackage{threeparttable}
\usepackage{subfigure}
\usepackage{url}
\usepackage{amsmath} 
\usepackage{amssymb}  
\usepackage{color}
\usepackage{xcolor}
\usepackage{float} 

\definecolor{skyblue}{rgb}{0.529, 0.808, 0.922} 
\definecolor{lightblue}{rgb}{0.678, 0.847, 0.902}
\usepackage[ruled,vlined]{algorithm2e}
\usepackage{algpseudocode}
\usepackage{gensymb}
\usepackage{bigstrut,multirow}
	
\usepackage[colorlinks=false, pdfborder={0 0 0}]{hyperref}
\usepackage{cleveref}
\usepackage{booktabs} 
\usepackage{multirow}
\usepackage{graphicx}
\usepackage{makecell}

\crefname{figure}{fig.}{figs.}
\crefname{table}{tab.}{tabs.}
\crefname{equation}{eq.}{eqs.}
\usepackage{microtype}

\title{\LARGE \bf
MR-COGraphs: Communication-efficient Multi-Robot Open-vocabulary Mapping System via 3D Scene Graphs
}

\author{Qiuyi Gu$^{1*}$, Zhaocheng Ye$^{1*}$, Jincheng Yu$^{1}$, Jiahao Tang$^{1}$, Tinghao Yi$^{23}$,  Yuhan Dong$^{1}$, Jian Wang$^{1}$, \\ Jinqiang Cui$^{4}$, Xinlei Chen$^{1}$, Yu Wang$^{1}$  
\thanks{$^{1}$ Tsinghua University, Beijing, China.}%
\thanks{$^{2}$ Efort Intelligent Equipment, Wuhu, Anhui, China.}%
\thanks{$^{3}$ University of Science and Technology of China, Hefei, Anhui, China.}%
\thanks{$^{4}$ Pengcheng Laboratory, Shenzhen, Guangdong, China.}%
\thanks{$^{*}$ Contributed equally to this work.}%
\thanks{
This research was supported by the National Natural Science Foundation of China (No.U19B2019, 62203257, M-0248), Tsinghua University Initiative Scientific Research Program, Tsinghua-Meituan Joint Institute for Digital Life, Beijing National Research Center for Information Science, Technology (BNRist), Beijing Innovation Center for Future Chips.
}
}

\begin{document}

\maketitle
\thispagestyle{empty}
\pagestyle{empty}

\begin{abstract}
Collaborative perception in unknown environments is crucial for multi-robot systems.
With the emergence of foundation models, robots can now not only perceive geometric information but also achieve open-vocabulary scene understanding.
However, existing map representations that support open-vocabulary queries often involve large data volumes, which becomes a bottleneck for multi-robot transmission in communication-limited environments. 
To address this challenge, we develop a method to construct a graph-structured 3D representation called COGraph, where nodes represent objects with semantic features and edges capture their spatial \textcolor{black}{adjacency} relationships.
Before transmission, a data-driven feature encoder is applied to compress the feature dimensions of the COGraph.
Upon receiving COGraphs from other robots, the semantic features of each node are recovered using a decoder. 
We also propose a feature-based approach for place recognition and translation estimation, enabling the merging of local COGraphs into a unified global map.
We validate our framework \textcolor{black}{on two realistic datasets and the real-world environment.}
The results demonstrate that, compared to \textcolor{black}{existing baselines for open-vocabulary map construction,
our framework reduces the data volume by over 80\% while maintaining mapping and query performance without compromise.}
For more details, please visit our website at 
\url{https://github.com/efc-robot/MR-COGraphs}.
\end{abstract}

\section{Introduction}
\label{sec:intro}
Multi-robot systems have emerged as powerful solutions for perception in large unknown environments \cite{corah2019communication}.
The primary advantage of such systems lies in their ability to leverage distributed sensing and computing, enabling robots to share information and shorten task completion time. 
As the scale of these systems grows, the need to share environmental information to maintain system operations increases, necessitating data-efficient map representations for transmission.

\begin{figure}[t]
    \centering
    \includegraphics[width=1\linewidth]{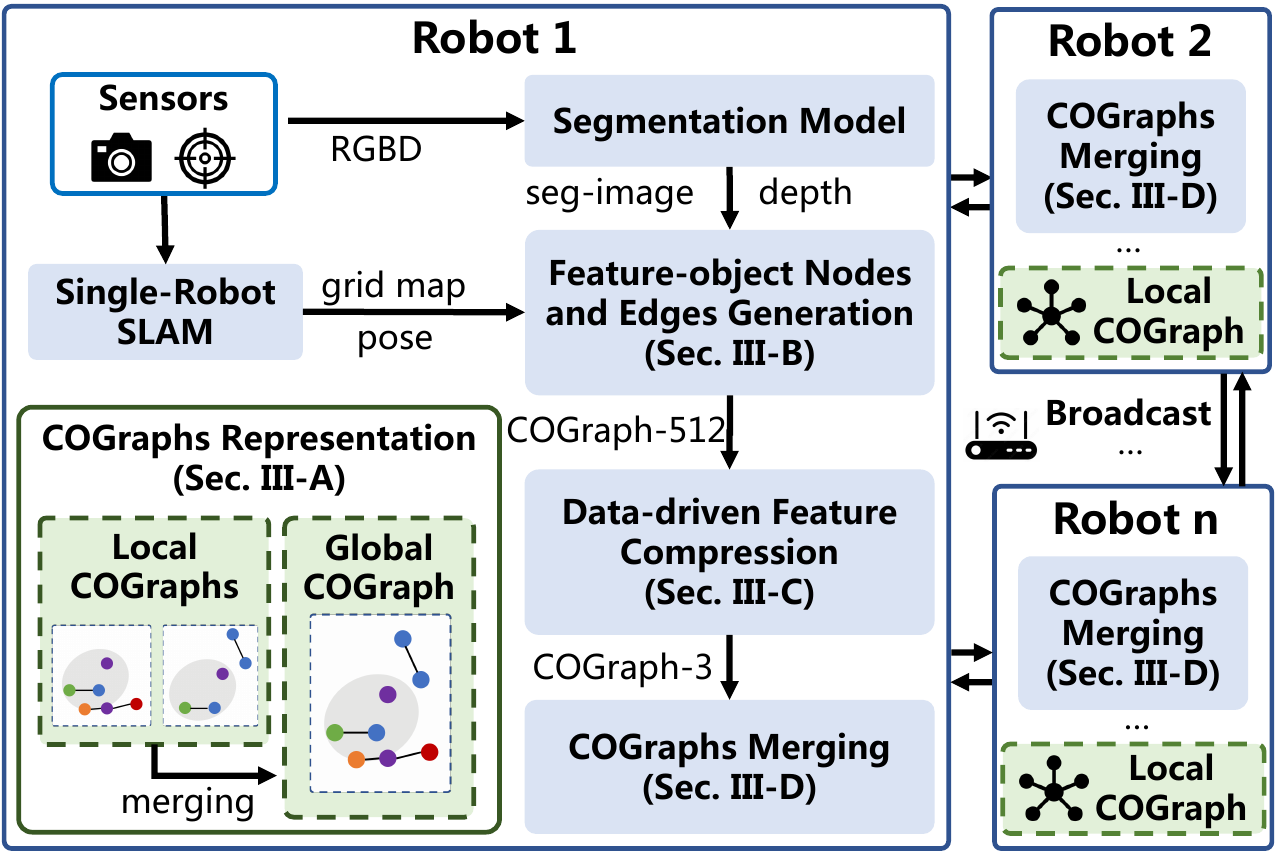}
    \caption{\textcolor{black}{Overview of the MR-COGraphs Framework.} 
    }
    \label{fig:sysframe}
    \vspace{-4mm}
\end{figure}

Recent advances in visual foundation models (\textit{e.g.}, SAM \cite{kirillov2023segment}) and vision-language models (\textit{e.g.}, CLIP \cite{radford2021learning}) have enabled the development of open-vocabulary 3D map representations. 
Traditional semantic maps \cite{grinvald2019volumetric} \cite{rosinol2020kimera} rely on predefined labels to describe the semantic information of the environment.
They are closed-vocabulary since their labels are confined to the classes of objects annotated in the training datasets \cite{he2017mask}.
In contrast, open-vocabulary maps are not constrained by predefined classes and can understand new categories or words without retraining. 
This is often achieved by extracting semantic feature vectors from images and projecting them into the 3D space \cite{jatavallabhula2023conceptfusion}. 
Due to their strong semantic understanding ability and adaptability, open-vocabulary representations unlock new possibilities for various language-guided tasks such as object retrieval \cite{peng2023openscene}, language-based navigation \cite{gu2024conceptgraphs} and manipulation \cite{shen2023distilled}.
 
However, current open-vocabulary 3D map representations demand significant data storage \textcolor{black}{\cite{gu2024conceptgraphs}\cite{werby2024hierarchical}}, which \textbf{becomes a communication bottleneck for multi-robot mapping systems.}
\textcolor{black}{For instance, in an open-vocabulary point cloud map, each point is linked to a high-dimensional feature vector \cite{gu2024conceptgraphs}. Consequently, the size of the map expands rapidly as the robot continuously perceives and records its surroundings.}
Even constructing a small-scale tabletop scene can require as much as 1.3GB of data storage \cite{lu2023ovir}.
This data explosion makes it difficult for multiple robots to share and update maps in real time.

3D scene graphs (3DSGs) are favorable for semantic mapping in communication-constrained environments due to their compact and flexible representation of the scene \cite{gu2024conceptgraphs}.
They model the environment as graph structures, with nodes representing every object's attributes and edges encoding the relationships between these objects.
Many studies have utilized \textcolor{black}{3DSGs} for large-scale semantic mapping \cite{hughes2022hydra} \cite{chang2023hydra}, hierarchical 3D scene construction \cite{armeni20193d} \cite{chen2023not}, and robot task planning \cite{gu2024conceptgraphs} \cite{rana2023sayplan}.
However, these approaches are largely focused on single-robot systems.
Although a few multi-robot mapping works \cite{chang2023hydra} \cite{chang2023d} have explored the collaborative construction of 3D scene graphs, \textbf{they do not consider open-vocabulary capabilities and have yet to address the critical challenge of reducing data size for efficient communication}.

To fulfill the requirements above, we propose a \textbf{C}ommunication-efficient \textbf{M}ulti-\textbf{R}obot \textbf{O}pen-vocabulary 3D Scene \textbf{Graphs}-based Mapping (MR-COGraphs) System with the following contributions:
\begin{itemize}
\item
A data-efficient open-vocabulary 3D scene graph construction method, in which a data-driven feature encoder compresses the dimension of features in COGraphs without losing semantic information.

\item 
A communication-efficient distributed multi-robot mapping system,  leveraging the semantic features of local COGraphs shared among robots to achieve place recognition and translation estimation.

\item 
We build and open source both simulated and real-world datasets to evaluate the performance of our system. 
Compared with baselines, our framework can reduce 80.11\%-89.80\% of data volume and our feature compression strategy reduces communication overhead by 95.72\%.
\end{itemize}

As illustrated in \Cref{fig:sysframe}, we propose a graph-structured open-vocabulary representation called COGraph (detailed in \Cref{sec:3A}).
Firstly, each robot generates the nodes and edges of its local COGraph utilizing the output of the Simultaneous Localization and Mapping (SLAM) module and the segmentation model (detailed in \Cref{sec:3B}).
Then a data-driven lightweight feature encoder (detailed in \Cref{sec:3C}) is employed to resize the 512-dimensional semantic features of nodes into 3 dimensions.
When receiving local COGraphs from other robots, the robot performs place recognition and translation estimation to merge local COGraphs into a global COGraph (detailed in \Cref{sec:3D}).
Afterward, experimental results are presented in \Cref{sec:exp}.
\Cref{sec:con} concludes this work and suggests future research directions.

\section{Related work}
\label{related work}

\subsection{3D Scene Graphs}

The concept of 3D scene graphs is first introduced in Armeni et al. \cite{armeni20193d}, where the authors propose a semi-manual method to extract buildings, rooms, objects, and cameras from the environment, creating a multi-layer graph structure. 
This approach abstracts the environment into nodes and edges, where each node can encompass multiple attributes, and edges represent spatial relationships and hierarchies.
Therefore, this representation is highly flexible, allowing for adjustments in its complexity and data volume.

There has been much progress in constructing 3DSGs with closed vocabulary \cite{hughes2022hydra} \cite{chen2023not} \cite{rosinol20203d}.
Hydra \cite{hughes2022hydra} leverages HRNet \cite{wang2020deep} as a pre-trained model to obtain semantic labels. It adds a mesh layer to enable the online generation of room and building nodes.
In StructNav \cite{chen2023not}, a robot employs visual SLAM along with MaskRCNN \cite{he2017mask} to develop a structured representation, and semantics are then integrated into geometry-based frontiers to facilitate object-goal navigation.

For open-vocabulary 3D scene graphs, OVSG \cite{chang2023context} presents an offline method to build nodes and edges based on OVIR-3D \cite{lu2023ovir}. 
\textcolor{black}{ConceptGraphs \cite{gu2024conceptgraphs} and HOV-SG \cite{werby2024hierarchical} are created by fusing the 2D outputs of foundation models into 3D space using collected RGBD sequences and poses. Although various language-guided planning tasks are presented to demonstrate their utility, they are computationally intensive and do not support online construction.} 
Clio \cite{maggio2024clio}, \textcolor{black}{a concurrent work to ours,} utilizes the information bottleneck principle to evaluate task relevance and proposes an online framework to construct task-driven scene graphs with embedded open-set semantics. 
In this work, we adopt an approach similar to \cite{gu2024conceptgraphs} and \cite{maggio2024clio}, with an added focus on further reducing the size of the scene graphs.
\textbf{The aforementioned approaches are single-robot frameworks} and methods for multi-robot 3DSGs construction are introduced in \Cref{sec:Multi-robot Mapping System}.

\subsection{Multi-robot Mapping System}
\label{sec:Multi-robot Mapping System}
In communication-limited environments, it is crucial to minimize the data transmission between robots while ensuring cooperative mapping and relative pose estimation.
To achieve this, SMMR-Explore \cite{yu2021smmr} only transmits submaps in the form of 2D occupancy grids and reconstructs place descriptors and point clouds locally.
Building on this, MR-TopoMap \cite{zhang2022mr} and MR-GMMExplore \cite{wu2022mr} further reduce data volume by transmitting topological maps and Gaussian Mixture Model (GMM) maps.
However, these approaches \textbf{only focus on constructing geometric maps and utilizing geometric features for map merging.}

Existing multi-robot semantic mapping systems are closed vocabulary \cite{chang2023hydra} \cite{chang2023d} \cite{chang2021kimera} \cite{fernandez2024multi}. 
Kimera-Multi \cite{chang2021kimera} implements a distributed semantic-metric SLAM system to construct the 3D mesh model of the environment.
Hydra-multi \cite{chang2023hydra} extends Hydra \cite{hughes2022hydra} into a centralized multi-robot system where each robot publishes its entire local scene graph including a 3D mesh layer and a control station handles relative transform estimation.
D-Lite \cite{chang2023d} addresses the communication constraint in navigation-oriented multi-robot coordination. It employs graph theory to compress 3DSGs by greedily preserving the shortest paths between locations of interest.
\textcolor{black}{Multi S-graphs \cite{fernandez2024multi} is a LiDAR-based multi-robot scene graph construction framework. It utilizes the semantic information embedded in situational graphs for cooperative map generation where only room-based descriptors and walls are transmitted for loop closures.}

\begin{figure*}[t]
    \centering
    \includegraphics[width=1\linewidth]{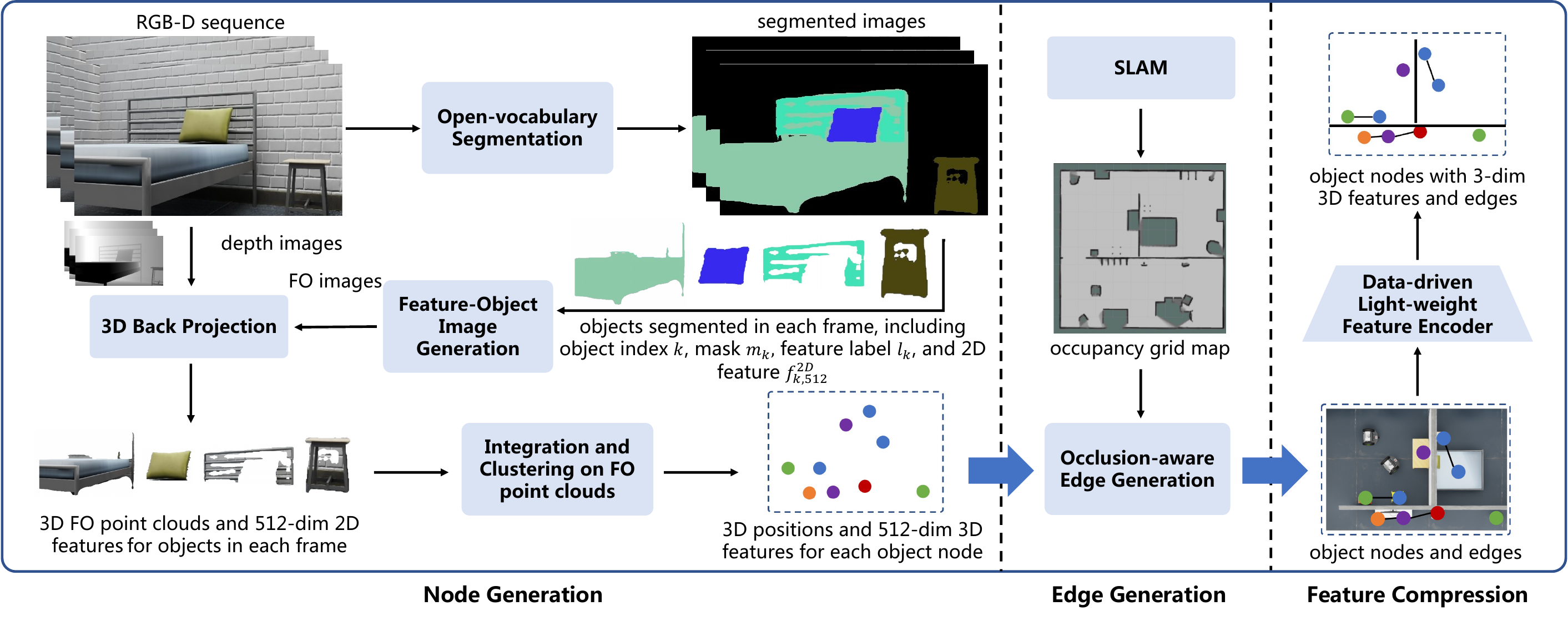}
    \setlength{\abovecaptionskip}{-0.5cm} 
    \setlength{\belowcaptionskip}{-0.4cm}
    \caption{The Generation Process of COGraphs.
    }
    \label{fig:genration}
    \vspace{-1mm}
\end{figure*}

\subsection{Open-vocabulary Scene Understanding}
With the emergence of foundation models, many works have begun to study language-guided object retrieval \textcolor{black}{\cite{peng2023openscene} \cite{lu2023ovir}}, which aims to locate objects based on text queries using open-vocabulary scene representations.
Explicit representations can update maps incrementally, making them more suitable for multi-robot systems. Therefore, we focus on open-vocabulary explicit representations.
They are typically achieved by projecting 2D semantic features onto 3D points and then encoding features into points \cite{jatavallabhula2023conceptfusion} \cite{peng2023openscene}, instances \cite{lu2023ovir} \cite{takmaz2023openmask3d}, and Gaussians \cite{qin2024langsplat} \cite{zhou2024feature}.

In OVIR-3D \cite{lu2023ovir} and OpenMask3D \cite{takmaz2023openmask3d}, the masks of instances are generated using foundation models and prior point clouds, and then the 3D features of each instance are computed offline.
ConceptFusion \cite{jatavallabhula2023conceptfusion} integrates pixel-aligned open-vocabulary features into 3D point cloud maps by combining traditional SLAM with multi-view images.
Similarly, OpenScene \cite{peng2023openscene} relies on prior point clouds and stores semantic features for each pixel, \textbf{resulting in a large data volume.}

3D Gaussian Splatting provides another way to realize open-vocabulary scene understanding using 3D Gaussian point cloud techniques.
Recent studies \cite{qin2024langsplat} \cite{zhou2024feature} integrate pre-trained 2D semantic features into 3D Gaussians to train a semantic 3DGS model, enabling object retrieval on rendered images.
Notably, LangSplat \cite{qin2024langsplat} introduces a neural network to reduce the dimensionality of CLIP features, thereby accelerating the training process.
Our work proposes an encoder-decoder strategy to compress semantic features for efficient data transmission.

\section{Method}
\label{sec:alg}
\textcolor{black}{
As shown in \Cref{fig:sysframe}, this section first outlines the map representation of the COGraph, followed by an introduction to the three key modules:
1) feature-object nodes and edges generation, which constructs the COGraph with high-dimensional features;
2) data-driven feature compression, where node features are encoded into three dimensions;
3) COGraphs merging, where local COGraphs from different robots are integrated into a unified COGraph.
}

\subsection{COGraphs Representation}
\label{sec:3A}
The proposed COGraph consists of the robot name, nodes, and edges.  
Each node contains the information presented in \Cref{table:node} while each edge only contains the information to identify the adjacency between nodes, including the robot name and IDs of the two nodes.
To reduce data transmission, the 512-dimensional features of each node are stored locally while only the 3-dimensional features along with other information in the COGraphs are transmitted. 
Upon receiving COGraphs from other robots, the 512-dimensional features are reconstructed from the 3-dimensional features using a feature decoder. 
Additionally, only newly generated nodes and edges are shared during communication.

\begin{table}[h]
    \caption{The Information One Node Contains in COGraphs}
    \centering
    \begin{tabular}{l l l l}
    \toprule
    \textbf{Symbol} & \textbf{Description} & \textbf{Size} & \textbf{Transmit} \\
    \midrule
    $N$        & robot name         & 8 bits  & yes         \\
    $i$        & node ID    & 8 bits   & yes         \\
    $pos_{i}$ & 3D center position  & 96 bits & yes         \\
    $l_{i}$    & feature label  & \textcolor{black}{16 bits} & yes         \\
    $b_{i}$    & bounding box  & 24 bits  & yes         \\
    $f_{i,512}^{3D}$   & 512-dimensional feature  & 4096 bits & no         \\
    $f_{i,3}^{3D}$     & 3-dimensional feature & 24 bits & yes         \\
    \bottomrule
    \end{tabular}
    \label{table:node}
\end{table}

\subsection{Feature-object (FO) Nodes and Edges Generation}
\label{sec:3B}
As illustrated in \Cref{fig:genration}, given a sequence of RGB-D images, we run an open-vocabulary segmentation model to obtain the segmented objects in each frame. 
\textcolor{black}{
Since the segmentation model is instance-aware, each object has an object index $k$, a 2D mask $m_k$, a feature label $l_k$ , and a 2D feature $f_{k,512}^{2D}$.
As a result, each object is assigned a unique 512-dimensional feature while multiple objects may share the same feature label due to semantic similarities.
We design a feature-object (FO) encoding strategy that generates FO images for each frame to facilitate precise object differentiation and node generation.
Specifically, each pixel in the FO images is encoded with 24 bits: 16 bits represent the feature label $l_k$, and the remaining 8 bits store the object index $k$.
The feature label $l_k$ is then used for subsequent clustering to generate nodes.} These FO images can be transmitted in the same way as existing image formats, and various lossless image compression methods can also be employed.

3D back projection is conducted using FO images, depth images, and poses derived from SLAM. 
To ensure localization accuracy and robustness, we employ a LiDAR-based SLAM algorithm to estimate robot poses and generate occupancy grid maps.
Then the 3D FO point clouds $pc_k$ and the corresponding 2D feature $f_{k,512}^{2D}$ for each object $k$ in every frame are generated.
We integrate semantic point clouds from adjacent frames according to the first 16 bits of information of each point in $pc_k$.
Then clustering is performed using an approach similar to Hydra \cite{hughes2022hydra}.
The output is a set of nodes, each characterized by a node ID $i$, a center position $pos_{i}$, a feature label $l_{i}$, and a bounding box $b_{i}$.

A separate thread is performed to compute the 3D semantic features for each object.
We determine whether the semantic point clouds $pc_k$ in a sequence of frames belong to the same object by analyzing the information encoded in FO images.
If they do, and the overlap between these point clouds exceeds a predefined threshold, the 2D features $f_{k,512}^{2D}$ associated with these point clouds are averaged to produce 3D features $f_{k,512}^{3D}$. 
The corresponding point clouds are then merged.
Given each 512-dimensional 3D feature $f_{k,512}^{3D}$ and its corresponding point clouds, we assign it to the nearest node by selecting the one closest to the center of the merged semantic point cloud, resulting in the node feature $f_{i,512}^{3D}$.

Edges are created if: 1) the distance between nodes is below a threshold, and 2) no occlusion exists between them. While nodes include center positions and bounding boxes to describe spatial relationships, background elements like walls are often missed by the segmentation model due to their size and lack of texture. To address this, we use the 2D grid map from SLAM to incorporate background information. Candidate edges are first generated based on node positions, and then refined using the occupancy grid map to remove connections between nodes in different rooms.

\begin{figure}[t]
    \centering
    \includegraphics[width=1\linewidth]{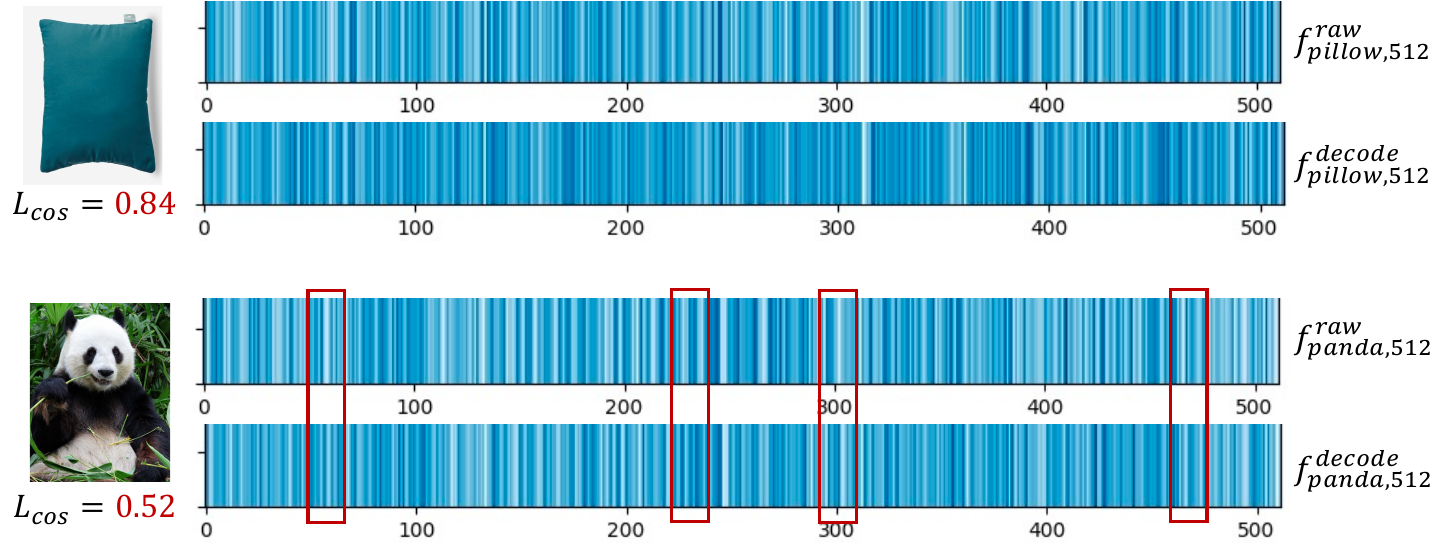}
    \setlength{\abovecaptionskip}{-0.4cm} 
    \setlength{\belowcaptionskip}{-0.4cm}
    \caption{\textcolor{black}{Comparison of the original and decoded features when the encoder and decoder are trained on household-related images from ImageNet.}
    }
    \label{fig:encoder}
    \vspace{-4mm}
\end{figure}

\subsection{Data-driven Feature Compression}
\label{sec:3C}
Compared to existing open-vocabulary 3D map representations, transmitting semantic nodes and edges generated in \Cref{sec:3B} significantly reduces data volume.
However, the 512-dimensional features of each node still pose communication overhead, particularly in bandwidth-limited environments. 
\textcolor{black}{To address this, we compress the features into 3 dimensions using a lightweight feature encoder. The original 512-dimensional features can later be recovered through a corresponding decoder for map merging and text query.}
\subsubsection{Model Structure}
The feature encoder is a multi-layer perceptron (MLP) that reduces dimensions from 512 to 3 through fully connected layers of sizes: 512, 256, 256, 128, 64, 32, 16, and 3. Each layer (except the first) is followed by BatchNorm1d and ReLU for normalization and activation.
Conversely, the feature decoder is an inverse MLP that expands dimensions from 3 back to 512. Its layer sizes are: 3, 8, 16, 32, 64, 128, 256, 256, and 512. Like the encoder, each layer (except the first) uses ReLU for non-linear transformation.

\subsubsection{Training Process}
\textcolor{black}{
We train the feature encoder and decoder using images from the ImageNet dataset \cite{imagenet-object-localization-challenge}, which contains over 80,000 images across 1,000 categories.
To accelerate training and enhance model reliability, we filter categories relevant to the target environment using a Large Language Model (LLM). 
The LLM is prompted with: ``Please find the categories that are related to [ ]," where [ ] specifies the environment type. 
From the selected categories, we extract corresponding images and their bounding boxes to obtain 512-dimensional CLIP image features. 
These features are then used to train the encoder and decoder, which are optimized to effectively compress and reconstruct high-dimensional features.
The loss function combines L2 loss and cosine similarity loss between the original feature $f_{i,512}^{raw}$ and the reconstructed 512-dimensional features $f_{i,512}^{decode}$, which is shown as:
\begin{equation}
\begin{aligned}
L = L_2 + L_{cos}
\label{equ:loss}
\end{aligned}
\end{equation}
where the cosine similarity loss is computed as:
\begin{equation}
\begin{aligned}
L_{cos} = \frac{f_{i,512}^{raw} \cdot f_{i,512}^{decode}}{|f_{i,512}^{raw}| \cdot |f_{i,512}^{decode}|}
\label{equa: similarity}
\end{aligned}
\end{equation}
}

\subsubsection{Feature Compression Process}
\textcolor{black}{We compress the semantic features of each node in the COGraph into a 3-dimensional representation using an encoder, as shown in \Cref{fig:genration}. 
However, in open-vocabulary environments, there are cases where the compressed features cannot be accurately reconstructed. 
For instance, as illustrated in \Cref{fig:encoder}, our trained encoder and decoder for ``indoor household scene" can accurately reconstruct features for objects like ``pillow" (with a cosine similarity $L_{cos} = 0.84$), but exhibit significant discrepancies for objects such as ``panda" (highlighted by the red box). 
Based on this observation, we conduct further experimental evaluations in \Cref{sec:exp2}.
Additionally, this issue can be mitigated through a pre-compression validation step based on the $L_{cos}$ value. 
Specifically, if $L_{cos}$ falls below a predefined threshold, the original high-dimensional features are retained without compression.
}

\begin{figure}[h]
    \centering
    \includegraphics[width=1\linewidth]{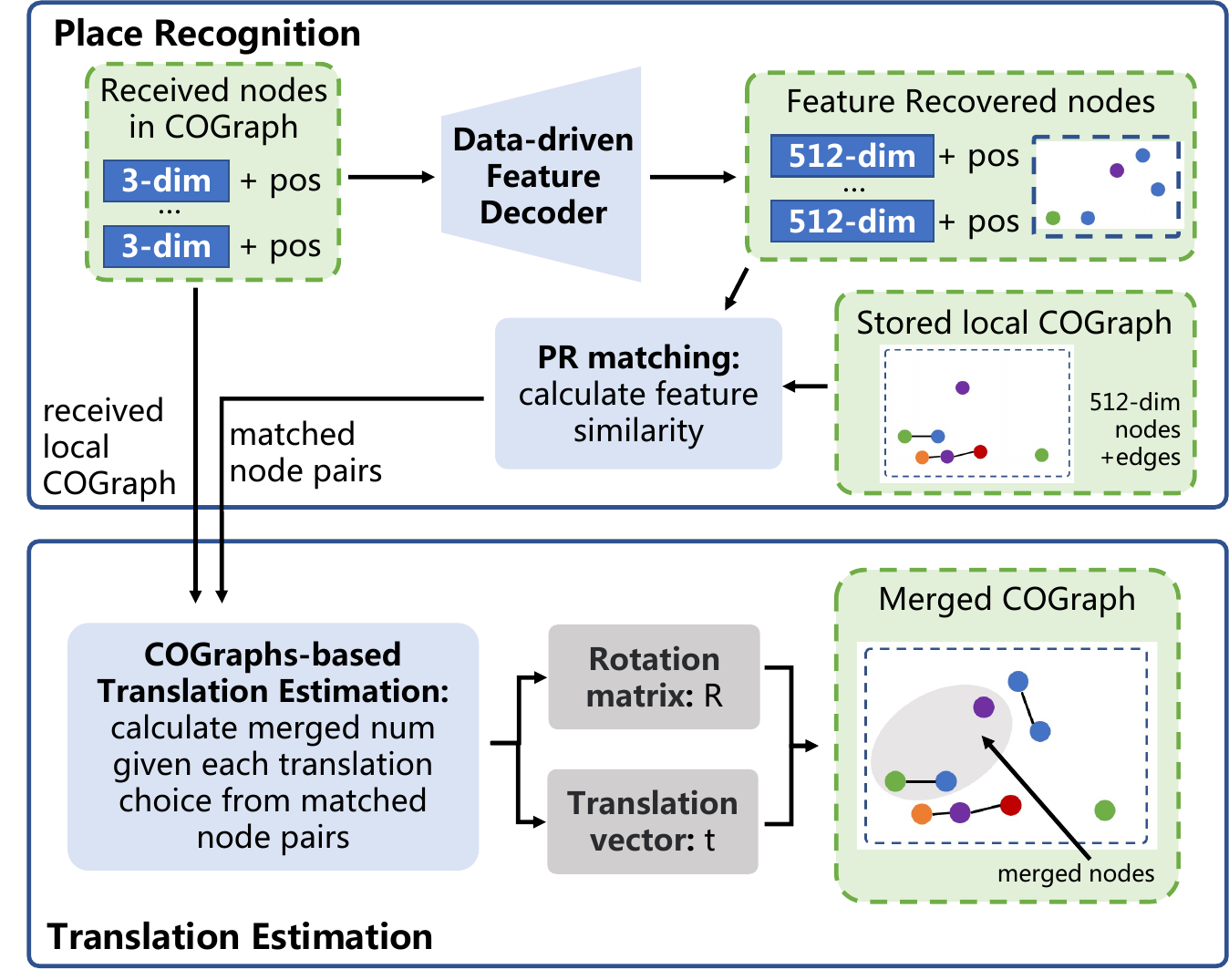}
    \setlength{\abovecaptionskip}{-0.5cm} 
    \setlength{\belowcaptionskip}{-0.4cm}
    \caption{COGraphs Merging.
    }
    \label{fig:merging}
    \vspace{-4mm}
\end{figure}

\subsection{COGraphs Merging}
\label{sec:3D}

When a robot receives COGraphs from other robots, it first determines whether they have passed the same area and then estimates their relative positions to merge the COGraphs. 
An illustration of the merging process is shown in \Cref{fig:merging}. 
\textcolor{black}{It is important to emphasize that our goal is not to construct a highly precise geometric map, but rather to collaboratively build a graph-structured map. As long as the merged COGraph is functional for downstream tasks, the precision of the merging method is not a critical requirement.}

\subsubsection{Place Recognition}
The feature decoder trained in \Cref{sec:3C} is used to recover each node's semantic feature $f_{i',512}^{3D}$ in received COGraphs.
Place recognition is then performed by iteratively calculating the feature similarity between each received node and nodes in the local COGraph.
When the similarity exceeds a predefined threshold, node $i'$ and node $i$ are marked as a matching pair.
If the number of matching pairs between the received and the local COGraph exceeds a threshold, it strongly indicates that the two robots have passed through the same area, prompting the translation estimation step.
\textcolor{black}{If there are not enough matching pairs, the COGraph received by the robot will remain local for future place recognition.}

\subsubsection{Translation Estimation}
Each robot establishes a local coordinate system with its starting position as the origin and its orientation as the X-axis. 
When merging maps from multiple robots, the rotation $R$ and translation $t$ between coordinate systems need to be estimated. 
Since the rotation can be directly obtained using a compass \cite{zhang2022mr}, only the translation vector $t$ needs to be estimated.
Two nodes from different robots are merged if their feature similarity and distance after coordinate transformation meet predefined thresholds.
We evaluate all candidate translations $t_{i,i'}$ from matching pairs and select the one that maximizes the number of merged nodes. 
\textcolor{black}{This translation selection method helps to avoid merging similar objects that appear in different locations.}
Finally, the merged COGraph is generated using the chosen translation vector $t$ and the rotation matrix $R$.


\section{Experiments}
\label{sec:exp}

\setlength{\tabcolsep}{4pt}
\begin{table}[htbp]
\centering
\caption{\textcolor{black}{Open-vocabulary 3D Scene Graphs Evaluation}}
\begin{tabular}{llcccccc}
\toprule
Scene & Method & $R_{obj}$ & R@1  & R@5 &  TPF(s) & Size(MB) \\
\midrule
\multirow{4}{*}{\makecell[c]{\rotatebox{90}{room0}}} & ConceptGraphs & 85.7 & 66.7 & 76.2 & 9.0 & 15.8 \\
                        & HOV-SG  & 76.2 & 47.6 & 71.4 & 48.0 & 141.0\\
                        & COGraph-512 & 85.7 & 42.9 & 76.2 & 0.145 & 0.0202\\
                        & COGraphs(ours) & 85.7 & 57.1 & 66.7 & 0.140 & 0.00084\\
\midrule
\multirow{4}{*}{\makecell[c]{\rotatebox{90}{office2}}} & ConceptGraphs & 70.0 & 45.0 & 60.0 & 12.6 & 9.8\\
                        & HOV-SG & 60.0 & 40.0 & 60.0 & 48.5 & 61.0\\
                        & COGraph-512 & 60.0 & 45.0 & 60.0 & 0.148 & 0.0127\\
                        & COGraph(ours) & 60.0 & 40.0 & 55.0 & 0.142 & 0.00053\\
\midrule
\multirow{4}{*}{\makecell[c]{\rotatebox{90}{apartment2}}} & ConceptGraphs & 66.7 & 28.6 & 42.9 & 15.0 & 8.1\\
                        & HOV-SG  & 66.7 &  33.3 & 61.9 & 6.0 & 130.0\\
                        & COGraph-512 & 81.0 & 42.9 & 66.7 & 0.152 & 0.0313\\
                        & COGraph(ours) & 81.0 & 47.6 & 61.9 & 0.141 & 0.00129\\
\bottomrule
\end{tabular}
\label{table:sota}
\end{table}

\textcolor{black}{
In this section, we
1) conduct experimental evaluations comparing our approach with state-of-the-art methods (\Cref{sec:exp1}),
2) analyze the open-vocabulary capabilities and design insights of our data-driven feature encoder (\Cref{sec:exp2}),
3) evaluate the robustness of our COGraph merging method under localization errors (\Cref{sec:exp3}), and
4) validate the framework in real-world environments (\Cref{sec:exp4}).
}

\subsection{Open-vocabulary 3D Scene Graphs Evaluation}
\label{sec:exp1}
\subsubsection{Dataset}
The Replica dataset \cite{replica} has been widely used in studies related to 3D scene reconstruction and object retrieval. 
It comprises 18 indoor environments, from which we select three representative scenes (room0, office2, and apartment2) due to their substantial size and rich semantic diversity.
To facilitate mapping and query evaluation, we develop a ROS wrapper \footnote{\url{https://github.com/efc-robot/replica-ros-wrapper}} to extract RGB-D sequences and ground-truth poses from the dataset, transforming them into ROS bag files for seamless integration with our framework.

\subsubsection{Metrics}
\label{sec: metrics}
We evaluate the accuracy of 3D Scene Graphs using the object finding rate $R_{obj}$ \cite{hughes2022hydra}, which measures the proportion of object nodes successfully constructed in the scene relative to the total number of objects.
We also utilize the query success rate R@k defined in ConceptGraphs \cite{gu2024conceptgraphs}.
This metric evaluates object retrieval capability by considering the top-k most likely objects in 3D Scene Graphs, with the retrieval counted successful if the correct object is among them. 
\textcolor{black}{Additionally, we record the average runtime per frame (tpf), and the total map volume.}

\subsubsection{Baselines}

\textcolor{black}{
We compare our approach with ConceptGraphs \cite{gu2024conceptgraphs} and HOV-SG \cite{werby2024hierarchical}.
Since our text queries do not include complex negation or multi-step affordances, we run ConceptGraphs without GPT.
We also test COGraph-512, a variant of our method without feature compression.}
Detic \cite{zhou2022detecting} is utilized as the open-vocabulary segmentation model in our framework.
Experiments in \Cref{sec:exp} are conducted on a desktop PC equipped with an Intel I7-13700 CPU and an Nvidia RTX 4080 GPU.

\subsubsection{Results}
\textcolor{black}{
\Cref{table:sota} presents a comprehensive evaluation.
Firstly, although both ConceptGraphs and HOV-SG have addressed map size reduction in their respective studies, they still retain point cloud data and high-dimensional semantic features for each object. 
In contrast, COGraph-512 significantly reduces the data volume, and COGraphs further optimizes this by storing 3-dimensional features.
It is worth noting that our lightweight feature compression model has a size of only 1.61 MB, which is substantially smaller than the map sizes of baseline methods.
Second, the results demonstrate that our feature compression process does not compromise the object finding rate and query success rate across the three evaluated scenes. 
A more detailed analysis of this aspect is provided in \Cref{sec:exp2}. 
Compared to baseline methods, \textbf{our approach not only maintains high accuracy and query success rates but also ensures real-time performance in the mapping system.
Taking the size of the feature compression model into consideration, COGraph reduces 80.11\%-89.80\% of data volume.}
}

\subsection{Feature Compression Evaluation}
\label{sec:exp2}



\textcolor{black}{
In this section, we address two key questions: 
1) whether feature compression reduces the open-vocabulary expressiveness of semantic features, and
2) how to design a feature compression model to minimize accuracy loss.}

\begin{figure}[h]
    \centering
    \includegraphics[width=1\linewidth]{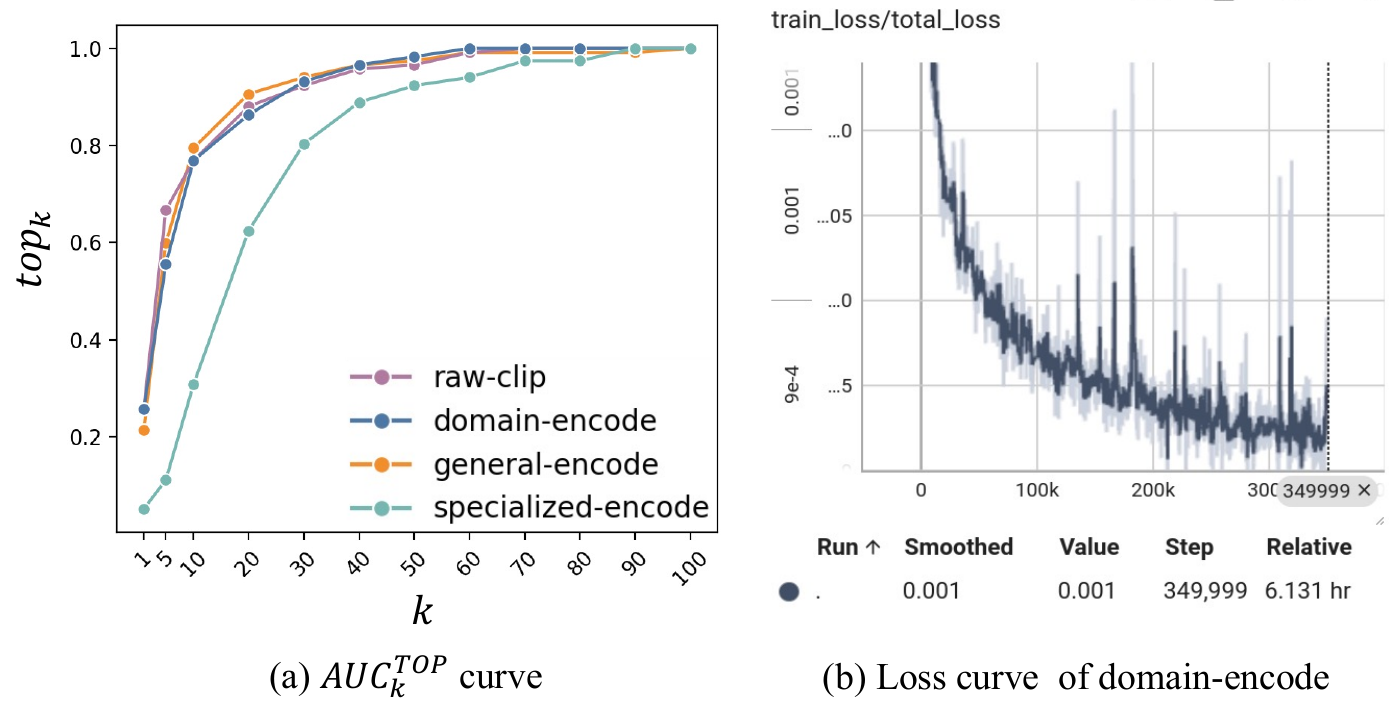}
    \setlength{\abovecaptionskip}{-0.5cm}
    \setlength{\belowcaptionskip}{-0.7cm}
    \caption{\textcolor{black}{Feature Compression Evaluation.}
    }
    \label{fig:AUC}
    \vspace{-2mm}
\end{figure}

\textcolor{black}{
We train the feature encoder and decoder using ImageNet images under three configurations:
1) \textbf{general-encode}: trained on all images,
2) \textbf{domain-encode} (used in our framework): trained on household-related images selected by prompting the LLM \cite{kimi_gpt} with ``household scene," and
3) \textbf{specialized-encode}: trained on animal and plant-related images selected by prompting the LLM with ``animals and plants."
The number of images LLM selected in the above configurations are 80477, 27284, and 21361.
The training parameters are set as follows: epochs = 5000, batch size = 1920, and learning rate = 0.0001. The loss curve for training domain-encode is visualized in \Cref{fig:AUC}b.}

\textcolor{black}{To comprehensively evaluate open-vocabulary semantics, we adopt the $AUC_{k}^{TOP}$ curve metric introduced in HOV-SG \cite{werby2024hierarchical}. 
As shown in \Cref{fig:AUC}a, the closer the $AUC_{k}^{TOP}$ curve is to the top left corner of the figure, the higher the open-vocabulary ability of the features.
To compute this metric, we evaluate the encoder and decoder on the Replica dataset by collecting 568 images annotated with 100 semantic categories. 
The $top_k$ score is calculated by querying all the 100 semantic texts and averaging the success rate that each image's true annotation is among the top-k semantic texts.}

\begin{figure*}[t]
    \centering
    \includegraphics[width=1\linewidth]{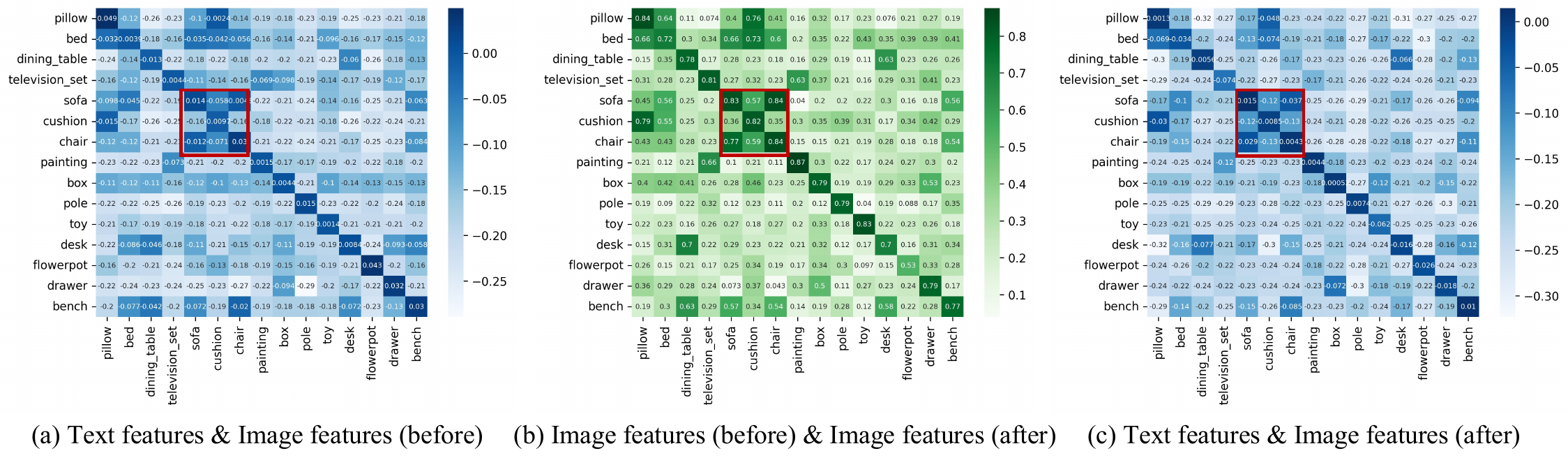}
    \setlength{\abovecaptionskip}{-0.5cm}
    \caption{Cosine Similarity between Text Features and Image Features (before feature encoding and after feature decoding).}
    \label{fig:featurecompression}
\end{figure*}

\textcolor{black}{As illustrated in \Cref{fig:AUC}a, we compare the performance of the three encoding configurations with \textbf{raw-clip}, which directly uses the 512-dimensional CLIP feature without encoding and decoding process.
The results show that both domain-encode and general-encode achieve performance comparable to \textbf{raw-clip}.
domain-encode performs the same with raw-clip when $k$ = 1 while the $top_1$ value of domain-encode is a bit lower.
This demonstrates that a well-designed encoder and decoder can preserve the open-vocabulary expressiveness of semantic features.
general-encode outperforms domain-encode when $k$ = 10 and 20.
However, specialized-encode performs poorly, likely due to its exclusive training on animal and plant features, which are not well-suited for the household environment of the Replica dataset. 
These findings provide valuable insights for designing feature encoders and decoders: 
\textbf{for better generalization, general-encode is recommended, while for domain-specific scenarios, training with domain-specific data (e.g. domain-encode) can yield superior open-vocabulary performance.}}

In \Cref{fig:featurecompression}, we further evaluate object retrieval performance by conducting auto-correlation and cross-correlation matching on features from 15 common objects.
\Cref{fig:featurecompression}a shows the matching results between the original CLIP image features and text features, while \Cref{fig:featurecompression}c presents the results after applying our feature compression strategy. 
In both figures, the deep color along the diagonal indicates high similarity between text and image features for the same object, while the lighter colors in other areas reflect low similarity for different objects.
This demonstrates that, after the data-driven compression process, the image features retain their ability to effectively match with the corresponding text features.
\textcolor{black}{Interestingly, as highlighted by the red square in \Cref{fig:featurecompression}, although these objects can be accurately retrieved using the R@1 metric, the sofa, cushion, and chair exhibit semantic similarities. This is likely because sofas and chairs share functional similarities, and cushions are often on top of them.}

\subsection{Map Merging Evaluation}
\label{sec:exp3}
\subsubsection{Dataset}
Since the Replica dataset lacks multi-room scenes suitable for collaborative mapping \cite{maggio2024clio} (only apartment2 is available), we construct two additional simulation environments, Isaac Small and Isaac Large, using the NVIDIA Isaac Sim platform \cite{isaac-sim}. 
As shown in \Cref{fig: setup}a, Isaac Large comprises a living room, bedroom, kitchen, and other areas. 
We employ the NVIDIA Carter robot \cite{robot-assets},  equipped with an RGB-D camera, 2D LiDAR, and IMU. 
For Isaac environments, Cartographer \cite{hess2016real} is used for localization, while ground-truth (GT) poses are utilized in the Replica environment.

\subsubsection{Metrics}
\textcolor{black}{
Unlike multi-robot SLAM, our localization module relies on a ready-made SLAM algorithm, and the graph-structured map does not require high geometric precision. 
However, to evaluate the accuracy of map merging, we still employ average translation $t_{error}$ \cite{yu2021smmr} to verify the correctness of the merged map.
}
Additionally, we analyze the amount of data transmitted between robots to assess communication efficiency.

\subsubsection{Results}

\begin{table}[h]
\centering
\caption{Map Merging Evaluation}
\begin{tabular}{ccccc}
\toprule
Scene & Dimension & Pose & ${t_{error}}$(m) & Data (KB) \\
\midrule
\multirow{4}{*}{Isaac Large} & \multirow{2}{*}{512}  & GT & 0.286 & 54.97 \\
                              &                      & Carto & 0.559 & 55.49 \\
\cmidrule{2-5}
                              & \multirow{2}{*}{3}   & GT & 0.138 & 2.35 \\
                              &                      & Carto & 0.559 & 2.37 \\
\midrule
\multirow{2}{*}{Replica Apartment2} & 512                   & GT & 0.084 & 27.9 \\
\cmidrule{2-5}
                              & 3                     & GT & 0.084 & 1.19 \\
\midrule
\multirow{2}{*}{Real-world} & 512                    & GT & 0.213 & 10.3 \\
\cmidrule{2-5}
                              & 3                     & GT & 0.213 & 0.44 \\
\bottomrule
\end{tabular}
\label{table:merging}
\end{table}

The map merging results are presented in \Cref{table:merging}.
Compared to a merging approach without feature compression, the increase in translation estimation error is minimal. 
Furthermore, localization errors are identified as the primary factor contributing to inaccuracies in translation vector estimation. 
Through feature encoding, the data exchanged between robots is reduced by approximately 95.72\%.
\Cref{fig:featurecompression}b illustrates the matching results of the image features before and after applying the encoder and decoder.
The deep color along the diagonal indicates that features processed through the encoder-decoder pipeline retain a high similarity to the original features while maintaining distinctiveness from other features.
In summary, reducing feature dimensions during transmission has minimal impact on COGraphs merging. 
This demonstrates that our method effectively reduces communication data volume without compromising mapping performance. 
\Cref{fig: setup}c visualizes the detected COGraph nodes in the real-world environment.
Videos of the merging process across the four environments are available on our website.

\subsection{Real-world Performance Evaluation}
\label{sec:exp4}

\begin{figure}[h]
    \centering
    \includegraphics[width=0.9\linewidth]{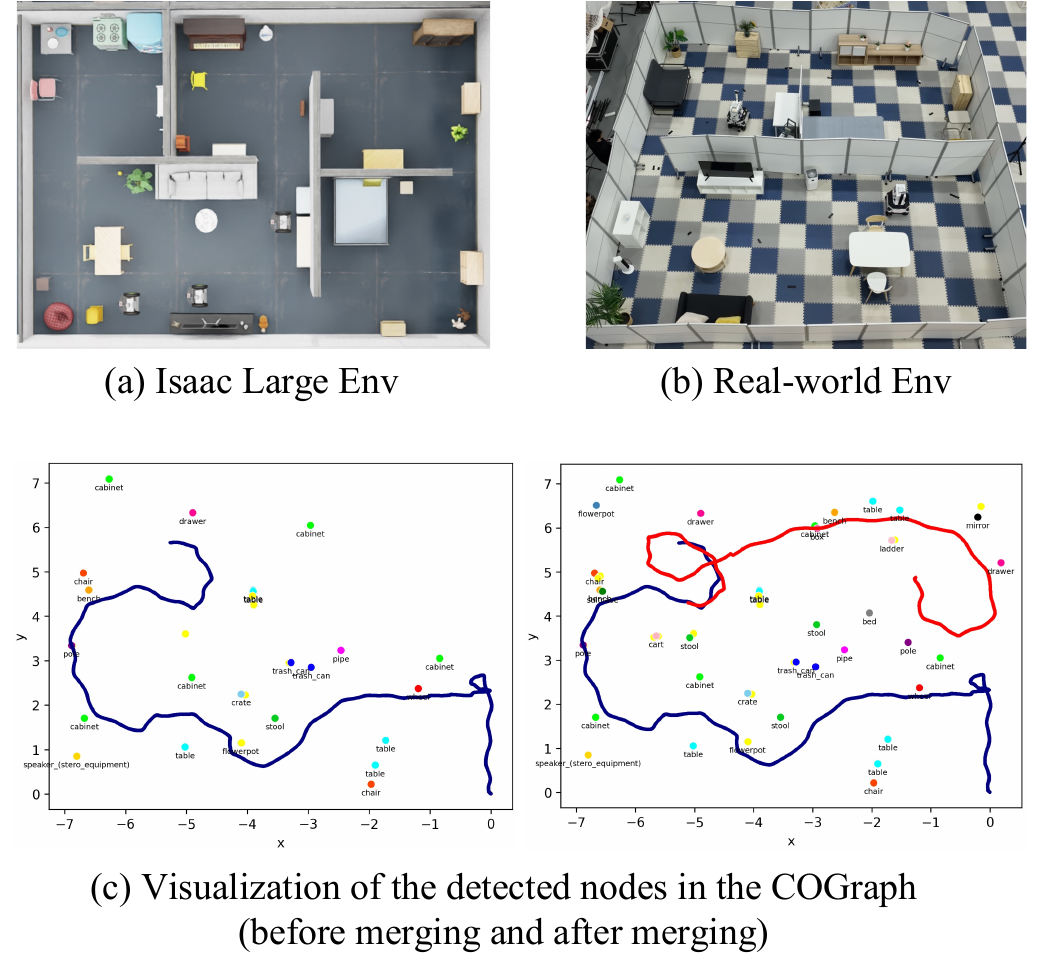}
    \caption{Experiment Environments and Visualization of the COGraph.
    }
    \label{fig: setup}
    \vspace{-2mm}
\end{figure}

As illustrated in \Cref{fig: setup}b, our real-world environment is 9m × 9m in size with 3 rooms.
Two robots equipped with iPhones are deployed for sensing.
We control them remotely to collect RGB-D images and pose information using an APP called Record3D \cite{record3d}.
Poses are derived using the built-in localization algorithm in Record3D and they are labeled as ``GT" for simplicity in \Cref{table:merging}.
\textcolor{black}{We have open-sourced a custom script that seamlessly converts sensed data into ROS topics, enabling real-time construction of COGraphs.}

\textcolor{black}{We run our framework on the desktop PC (mentioned in \Cref{sec:exp1}) and we also test it on the Nvidia Orin NX platform. 
The computation times for each module are detailed in \Cref{table:real-world}.
Additional real-world demonstrations are available on our website.}

\begin{table}[h]
\centering
\caption{Real-world Evaluation}
\begin{tabular}{ccccc}
\toprule
Platform & Segmentation & Mapping & Compression\\
\midrule
PC & 144 ms                    & 1428 ms & 0.174ms \\
\midrule
Orin NX & 1469 ms                    & 4250 ms & 2.126ms \\
\bottomrule
\end{tabular}
\label{table:real-world}
\end{table}

\section{Conclusion and Future Work}
\label{sec:con}
This paper presents a communication-efficient multi-robot mapping framework that allows for open-vocabulary object retrieval. 
We propose a method for constructing COGraphs, introducing a data-driven feature encoder to compress feature dimensions.
To facilitate multi-robot collaboration, we develop place recognition and translation estimation strategies based on semantic features for efficient COGraph merging.
Our framework is validated on realistic datasets and then real-world environment, demonstrating two orders of magnitude of reduction in data transmission while maintaining state-of-the-art 3DGS mapping accuracy and query success rates. 
In future work, we will leverage advanced vision-language models to further enhance open-vocabulary query capabilities and develop exploration strategies for autonomous semantic mapping in communication-constrained environments.

\bibliographystyle{IEEEtran}
\bibliography{6_reference}

\end{document}